\documentclass{article}
\usepackage{spconf,amsmath,graphicx}
\usepackage{amsmath,amsfonts}
\usepackage{multirow}
\usepackage{caption}
\usepackage{subcaption}
\usepackage{algorithm}
\usepackage{float}
\usepackage{graphicx}
\usepackage{tikz}
\usetikzlibrary{calc}
\usepackage{algpseudocode}
\usepackage{amsthm}

\usepackage{enumitem}
\usepackage{hyperref}
\usepackage{placeins}

\hypersetup{
  colorlinks=true,
  linkcolor=black,
  citecolor=black
}

\usetikzlibrary{arrows.meta, positioning, positioning, matrix, backgrounds, fit, external, plotmarks}

\theoremstyle{plain}
\newtheorem{definition}{Definition}[section]

\newcommand{\bk}{\mathbf{k}}
\newcommand{\bw}{\boldsymbol{\omega}}
\newcommand{\bs}{\mathbf{s}}
\newcommand{\bl}{\mathbf{q}}
\newcommand{\neigh}{\mathcal{N}_L}
\newcommand{\dd}{\mathrm{d}}
\newcommand{\R}{\mathbb{R}}
\newcommand{\Z}{\mathbb{Z}}
\newcommand{\be}{\mathbf{e}}

\usepackage{xcolor}
\definecolor{myblue}{RGB}{100, 150, 200}
\title{SplineSplat: 3D Ray Tracing for Higher-  Quality Tomography}
\name{Youssef Haouchat~$^{1}$, Sepand Kashani~$^{2}$, Aleix Boquet-Pujadas~$^{3}$, Philippe Thévenaz~$^{1}$, Michael Unser~$^{1}$}
\address{$^{1}$~Biomedical Imaging Group,
         $^{2}$~Center for Imaging -- EPFL, Lausanne, Switzerland\\
          $^{3}$~Paul Scherrer Institute (PSI) -- Villigen, Switzerland}

\begin{document}
\maketitle

\begin{abstract}
We propose a method to efficiently compute tomographic projections of a 3D volume represented by a linear combination of shifted B-splines. To do so, we propose a ray-tracing algorithm that computes 3D line integrals with arbitrary projection geometries. One of the components of our algorithm is a neural network that computes the contribution of the basis functions efficiently. In our experiments, we consider well-posed cases where the data are sufficient for accurate reconstruction without the need for regularization. We achieve higher reconstruction quality than traditional voxel-based methods. 
\end{abstract}


\section{Introduction}
Our 3D reconstruction scheme relies on an appropriate modification of a well-known ray-tracing algorithm: the digital differential analyzer (DDA) \cite{amanitides}. This extension enables the calculation of the x-ray transform of a continuous-domain representation of the volume which uses overlapping basis functions. We focus on well-posed settings where the measurements satify the Crowther criterion~\cite{rattey, Brooks1978} for a targeted resolution.
\subsection{State of the Art and Related Works}
The purpose of tomography is to reconstruct a 3D volume from a collection of line integrals which we refer to as projections. 
It is an inverse problem that is typically solved with optimization algorithms \cite{fessler_statistical_reconstruction} that require efficient implementations of the forward and adjoint operators. 
\begin{figure}[t]
\centering
\begin{tikzpicture}
  \node[anchor=south west, inner sep=0] (img) at (0,0)
    {\includegraphics[width=0.78\linewidth]{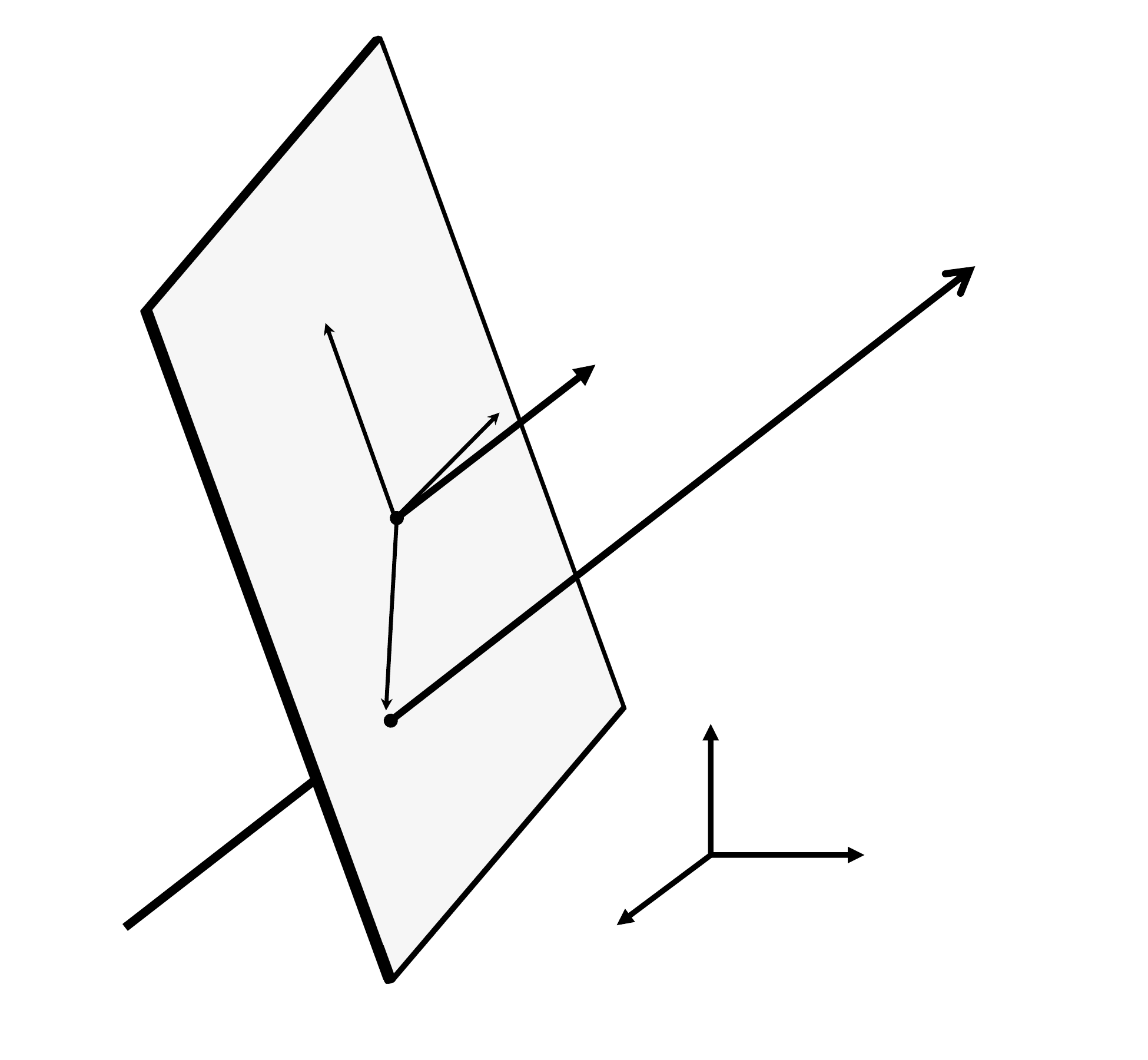}};

  \begin{scope}[x={(img.south east)}, y={(img.north west)}]

    --- Optional helper grid while positioning (turn on as needed) ---
    \node[anchor=west] at (0.62,0.31) {$\be_3$};
    \node[anchor=west] at (0.75 ,0.19) {$\be_2$};
    \node[anchor=east] at (0.6,0.1) {$\be_1$};

    \node[anchor=west] at (0.52,0.67) {$\bw$};
    \node[anchor=west, rotate=35] at (0.38,0.60) {$\mathbf{u}$};
    \node[anchor=east, rotate=25] at (0.34,0.71) {$\mathbf{v}$};
    \node[anchor=east] at (0.33,0.325) {$\bs$};

    \node[anchor=east] at (0.35,0.51) {$\mathbf{o}$};

    \node[anchor=west,align=left] at (0.68,0.8)
     {$\{\,\bs+t\bw,\ t\in\mathbb{R}\,\}$};

    \node[anchor=west, rotate=50] at (0.08,0.70) {\small $H_{\bw} = \mathrm{span}(\mathbf{u}, \mathbf{v})$};

  \end{scope}
\end{tikzpicture}
\caption{Notations for our parameterization of a line in 3D.}
\label{fig:ray_geometry}
\end{figure}
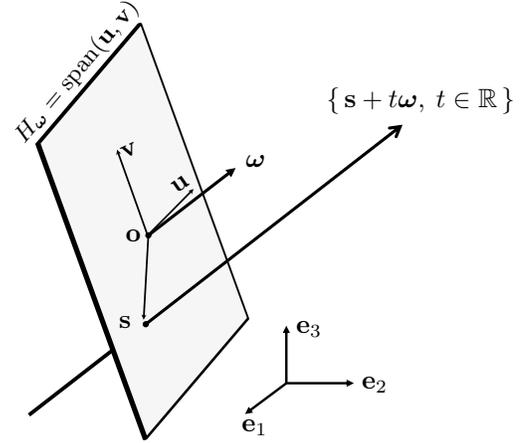

Ray-tracing algorithms are particularly well-suited for this purpose: They assume the volume is a piecewise-constant function defined over the cells---\emph{voxels}---of a grid. For each ray, they accumulate the intersected cell contributions by weighting the voxel values by the per-cell path lengths. This is often done with DDA, a broadly used algorithm in computer-graphics that effectively computes such intersections as the ray steps through the volume.

Efficient versions of DDA are available as open-source libraries such as the ASTRA toolbox~\cite{astra}, the TIGRE toolbox~\cite{tigre}, or the CONRAD framework~\cite{conrad}. The intrinsic limitations of these frameworks is their reliance on a piecewise-constant representation of the image (with voxels as basis functions). The voxel basis is known to be suboptimal for the approximation of smooth images~\cite{thevenaz_interpolation}. This ultimately affects the quality of the reconstructed images. To address this case, several works have proposed to use higher-order basis functions. The authors of \cite{mehrsa_pourya_box_splines} and \cite{entezari_unser} propose the use of box-splines in tomography, but their algorithm is computationally demanding and presently limited to 2D. Building on this idea, the work in \cite{haouchat_tci} generalizes the DDA algorithm for 2D images to support overlapping basis functions, and proposes closed-form expressions to account for their contributions in the ray-tracing routine.
In this paper, we extend this idea even further to 3D volumes and present a novel approach that overcomes the intractability of obtaining simple closed-form expressions for the x-ray projections of basis functions.

\subsection{Tomography and Discretization}
As in the setup in Figure~\ref{fig:ray_geometry}, we parameterize a line in $\mathbb{R}^3$ of direction \mbox{$\bw \in \mathbb{S}^2$} and offset $\bs \in {H}_{\bw}$ as
\begin{equation}
 \{\bs + t\bw, t \in \mathbb{R}\},
\end{equation}
where $\mathbb{S}^2 = \{\mathbf{x} \in \mathbb{R}^3, \|\mathbf{x}\|_2 = 1\}$ is the unit sphere and ${H}_{\bw} = \{\mathbf{x} \in \mathbb{R}^3, \langle \mathbf{x}, \bw \rangle = 0\}$ is the plane orthogonal to $\bw$.
The x-ray transform~\cite{natterer_mathstomo} $\mathcal{P}$ of an integrable function \mbox{$f:\mathbb{R}^3 \to \mathbb{R}$} corresponds to its integrals along such lines. It can be expressed in terms of $\bw$ and $\bs$ as
\begin{equation}
 \mathcal{P}f(\bw, \bs) = \int_{\mathbb{R}} f(\bs + t\bw) \dd t.
\end{equation}
In practice, the x-ray transform is computed on functions that are described by a finite number of parameters. A common approach is to represent $f$ as a linear combination of $N$ shifted versions of a single basis generator $\varphi:\mathbb{R}^3 \to \mathbb{R}$ as
\begin{equation}
  \label{eq:decomposition}
 f(\cdot) = \sum_{\bk \in \Omega} c_\bk \varphi(\cdot - \bk),
\end{equation}
where the $c_\bk \in \R$ are expansion coefficients and $\Omega~\subset~\mathbb{Z}^3$ is a finite index set. The linearity and shift property of the x-ray transform allows us to express $\mathcal{P}f$ as
\begin{equation}
  \label{eq:proj_decomp}
 \mathcal{P}f(\bw, \bs) = \sum_{\bk \in \Omega} c_\bk \mathcal{P}\varphi(\bw, \bs - \text{Proj}_{H_{\bw}} (\bk)),
\end{equation}
where $\text{Proj}_{H_{\bw}}(\cdot)$ is the orthogonal projection onto the plane $H_{\bw}$. The term 
$\mathcal{P}\varphi(\bw, \bs - \text{Proj}_{H_{\bw}} (\bk))$ is referred to as the \emph{contribution} of the basis function $\varphi(\cdot - \bk)$ to the value of the x-ray transform.
For a given ray, only basis functions that intersect its path contribute to \eqref{eq:proj_decomp}. The efficient evaluation of their indices $\bk$ and contributions $\mathcal{P}\varphi$ constitutes the main computational challenge.

\subsection{Contributions}
The computation of the 3D x-ray transform of a function that can be expressed as in \eqref{eq:decomposition} is challenging for two reasons. (1)~The basis functions overlap and extend over more than one cell. By contrast, DDA is designed for box-ray intersections, meaning contributions of off-centered cells are missed.  (2)~The contribution $\mathcal{P}\varphi$ of the basis generator to a given ray has to be computed explicitly. While closed-form expressions can be derived for specific basis functions in 2D, they are non-trivial to obtain in 3D. The contributions of the paper are as follows.
\begin{enumerate}[label=(\arabic*)]
  \item We propose a 3D geometry-agnostic algorithm that extends DDA to overlapping basis functions that are compactly supported.
  \item We introduce a learned component that computes the contribution of the basis function to a given line integral efficiently.
  \item We show that the use of higher-order basis functions leads to better reconstruction quality than traditional voxel-based methods in standard imaging scenarios where the inverse problem is sufficiently well-posed.
\end{enumerate}

\section{Method}
\subsection{Underlying Principles}
\label{sec:principles}
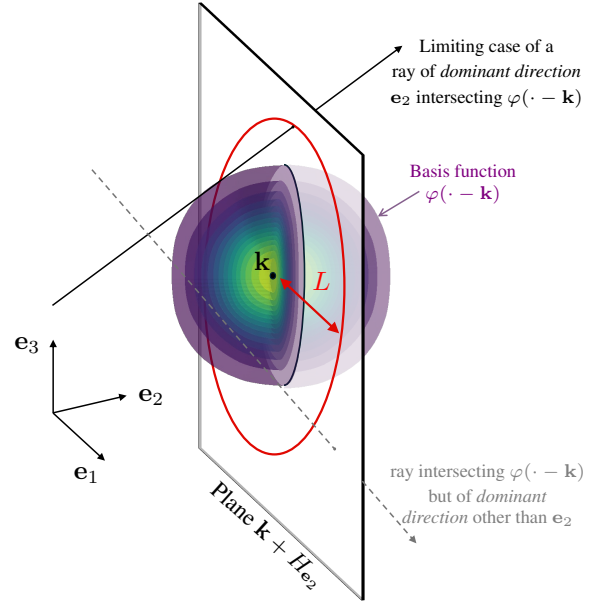
\begin{figure}[t]
\centering
\begin{tikzpicture}
  \node[anchor=south west, inner sep=0] (img) at (0,0)
    {\includegraphics[width=\linewidth]{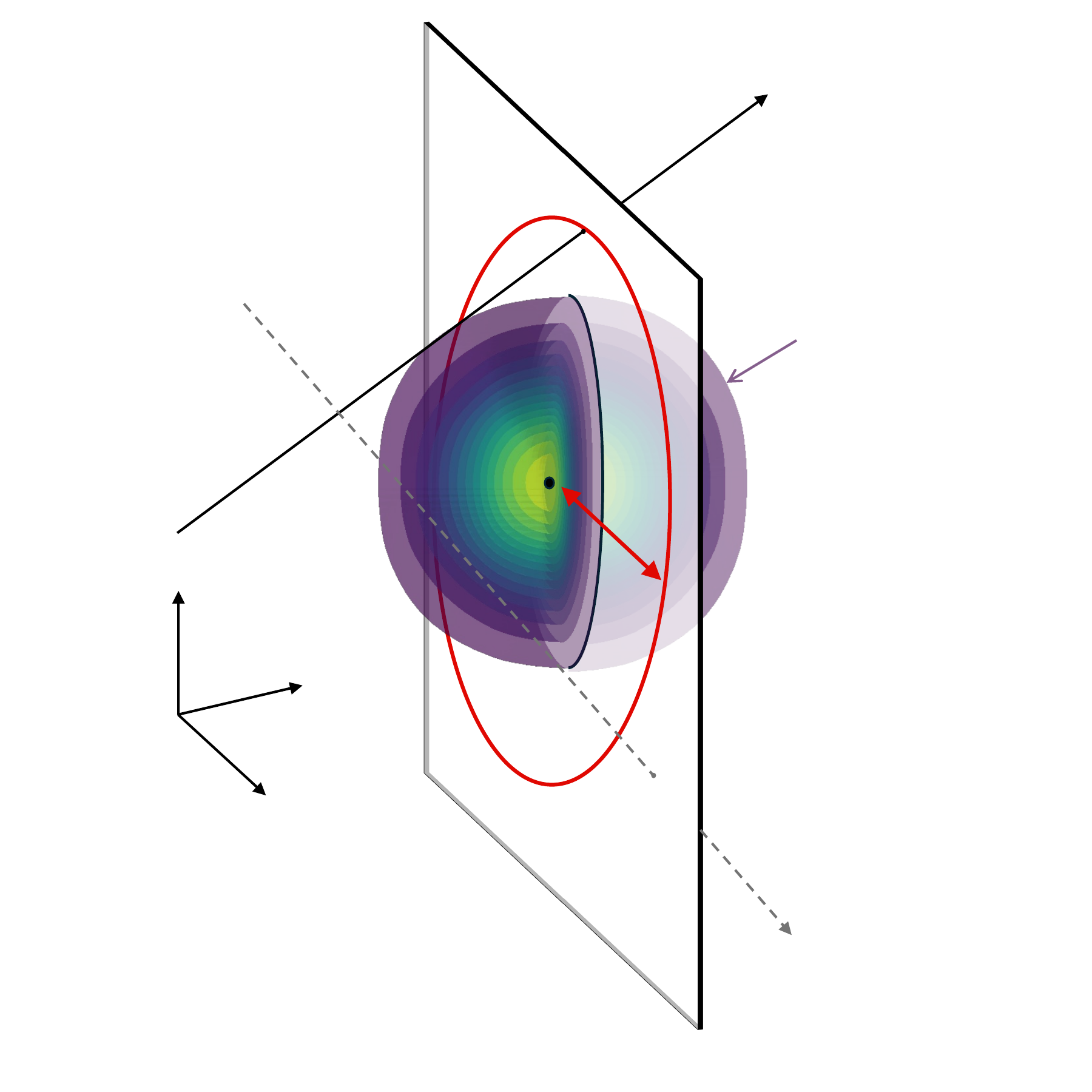}};

  \begin{scope}[x={(img.south east)}, y={(img.north west)}]

    --- Optional helper grid while positioning (turn on as needed) ---
    \node[anchor=west] at (0.09,0.45) {$\be_3$};
    \node[anchor=east] at (0.35,0.37) {$\be_2$};
    \node[anchor=east] at (0.25,0.25) {$\be_1$};
    \node[anchor=west, rotate=-42] at (0.4,0.24) {\small Plane $\bk + H_{\be_2}$};
    \node[anchor=west] at (0.55,0.55) {\color{red}$L$};
    \node[anchor=west] at (0.46,0.58) {$\bk$};
    \node[anchor=west] at (0.7,0.7) {\color{violet}\shortstack{\scriptsize Basis function  \\ \scriptsize $\varphi(\cdot - \bk)$}};
    \node[anchor=west] at (0.67,0.87) {\shortstack{\scriptsize Limiting case of a \\ \scriptsize ray of \emph{dominant direction} \\ \scriptsize $\be_2$ intersecting $\varphi(\cdot - \bk)$}};
    \node[anchor=west] at (0.67,0.22) {\color{gray}\shortstack{\scriptsize ray intersecting $\varphi(\cdot - \bk)$\\ \scriptsize  but of \emph{dominant}\\
    \scriptsize \emph{direction} other than $\be_2$}};
  \end{scope}
\end{tikzpicture}
\caption{Tensor-product quadratic B-spline basis function of index $\bk$ with \emph{maximal footprint radius} $L$.}
\label{fig:footprint}
\end{figure}

Let us introduce some terminology and notations. 
\begin{definition}[Dominant axis]
  Let $(\be_1, \be_2, \be_3)$
  be the canonical basis of $\R^3$. The \emph{dominant axis} of a direction $\bw=(\omega_1, \omega_2, \omega_3)
   \in \mathbb{S}^2$ is defined as an axis $\be_{d^*}$ such that
  \begin{equation}
    d^* \in \arg\max_{d=1,2,3} |\omega_d|.
  \end{equation}
\end{definition}

\begin{definition}[Maximal footprint radius]
  \label{def:footprint}
  Let \mbox{$\varphi:\R^3~\to~\R$} be a compactly-supported basis generator. We say that the \emph{maximal footprint radius} of $\varphi$ is $L$ if, $\forall d \in \{1,2,3\}$, all rays that have a direction of \emph{dominant axis} $\be_{d}$ that intersect the support of $\varphi$ also intersect the hyperplane $H_{\be_{d}}$ inside the disk of radius $L$ centered at the origin.
  Figure~\ref{fig:footprint} illustrates this. 
  \end{definition}

To simplify the notations, our basis functions are placed on a Cartesian lattice of stepsize 1. We consider a ray of direction $\bw$ with dominant axis $\be_{d^*}$. Let $\bk$ denote the index of the voxel that is currently crossed by the ray during DDA. We make two main observations that allow us to design an efficient ray-tracing algorithm for overlapping basis functions.

The first observation comes directly from Definition~\ref{def:footprint}.
It is that a basis function centered at $\bl \in \Z^3$ with \mbox{$\bl_{d^*} = \bk_{d^*}$} can have a nonzero contribution to the projection value only if \mbox{$\|\bl - \bk\|_\infty \leq \lceil L-\tfrac{1}{2} \rceil$}, where $L$ is the maximal footprint radius of the basis generator.
The upper bound with the \mbox{$\ell_\infty$-norm}, as well as the offset  $\tfrac{1}{2}$
come from the voxelized structure of the grid. By adding up the contributions of all such basis functions at each DDA step, we ensure that all basis functions that intersect the ray are accounted for.

The second observation is that, in DDA, the ray crosses at most three voxels on each plane orthogonal to the dominant axis, as shown in \cite{hao_gao}. Therefore we selectively exclude from the computation the basis functions that have already been accounted for in the two previous DDA steps, avoiding redundant computations.

\subsection{SplineSplat: Learning Spline Contributions}
\label{sec:splinesplat}
\begin{algorithm}[t]
\caption{Training of the {SplineSplat} network $f_\theta$}
\begin{algorithmic}[1]
\Require Discretized basis $\boldsymbol{\varphi}$ of maximal footprint radius $L$, MLP $f_\theta$, tolerance $\varepsilon$, learning rate $\eta$, batch size $B$
\vspace{0.1cm}
\While {$\tfrac{1}{N_\text{val}} \sum_{n=1}^{N_\text{val}} |y_n - f_{\theta}(\bw_n^\text{val}, \bs_n^\text{val})|^2 > \varepsilon$}
    \State $\bw_b \sim \text{Uniform}(\mathbb{S}^2)$ \quad $b=1,\ldots,B$
    \State $\bs_b \sim \text{Uniform}(H_{\bw_b}\cap \mathcal{B}(0, L))$ \quad $b=1,\ldots,B$
    \State Compute $y_b = \texttt{Project}(\boldsymbol{\varphi}, \bw_b, \bs_b)$ \quad $b=1,\ldots,B$    
    \State $\mathcal{L}(\theta) = \tfrac{1}{B} \sum_{b=1}^B |y_b - f_{\theta}(\bw_b, \bs_b)|^2$
    \State Update $\theta$ with Adam optimizer~\cite{kingma_adam}
    \EndWhile
    \State \Return $\theta$
\end{algorithmic}
\label{alg:training}
\end{algorithm}

Once the set of contributing basis functions is identified, their contributions to the projection value are estimated by a learned projector. We use a shallow multilayer perceptron (MLP)~$f_\theta$ to model the projector $\mathcal{P}\varphi$, where~$\theta$ denotes its learnable parameters. DDA is memory-bound, hence replacing closed-form expressions with shallow MLPs to compute $\mathcal{P}\varphi$ has no significant impact on runtime. The network inputs are a direction vector~$\bw\in\mathbb{S}^2$ and an offset~$\bs \in H_{\bw}$. To train the network, we sample random directions and offsets to minimize the~$\ell_2$ loss between the network output and ground-truth projections. The latter are obtained by sampling $\varphi$ on a much finer 3D grid and using a standard projector to evaluate $\mathcal{P}\varphi$. This is denoted by the \texttt{{Project}} function in Algorithm~\ref{alg:training} where $\boldsymbol{\varphi}$ denotes the dense samples of $\varphi$.

In our implementation,~$\varphi$ is a tensor-product quadratic B-spline, which is both compactly supported and smooth. The use of such basis functions has a theoretical justification provided by sampling theorems. As shown in~\cite{natterer_mathstomo}, a compactly supported function~$f \in L_1(\R^3)$ that is~$\varepsilon$-essentially bandlimited\footnote{
There exists $R > 0$ such that~$\int_{\|\boldsymbol{\xi}\|_2 > R} |\hat{f}(\boldsymbol{\xi})| \dd \boldsymbol{\xi} < \varepsilon \|f\|_{L_1}$.}can be reconstructed accurately from sufficiently dense x-ray samples. The sampling density directly depends on the essential bandwidth of~$f$. The spectrum of a square function (voxel) decays like $1/\|\boldsymbol{\xi}\|$, whereas that of a quadratic B-spline decays like $1/\|\boldsymbol{\xi}\|^3$. This implies that higher-order splines have a much smaller essential bandwidth, which ultimately leads to better reconstruction results for a fixed number of projections.

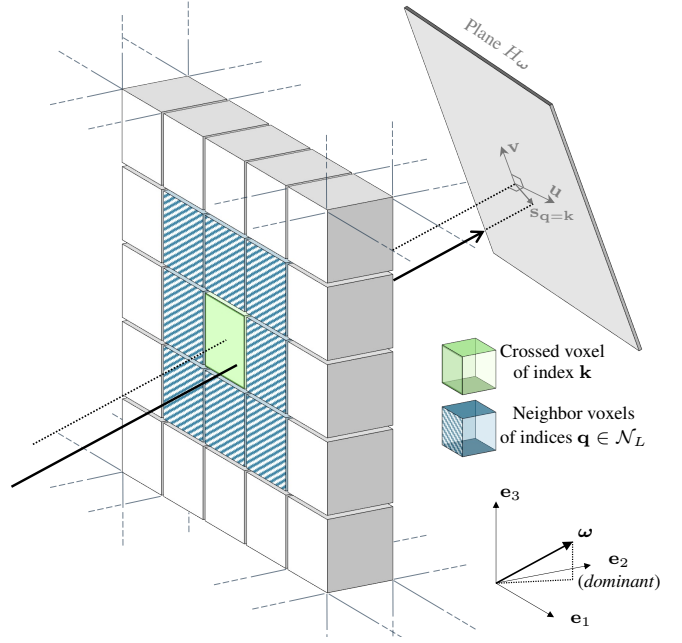
\begin{figure}
\centering
\begin{tikzpicture}
  \node[anchor=south west, inner sep=0] (img) at (0,0)
    {\includegraphics[width=\linewidth]{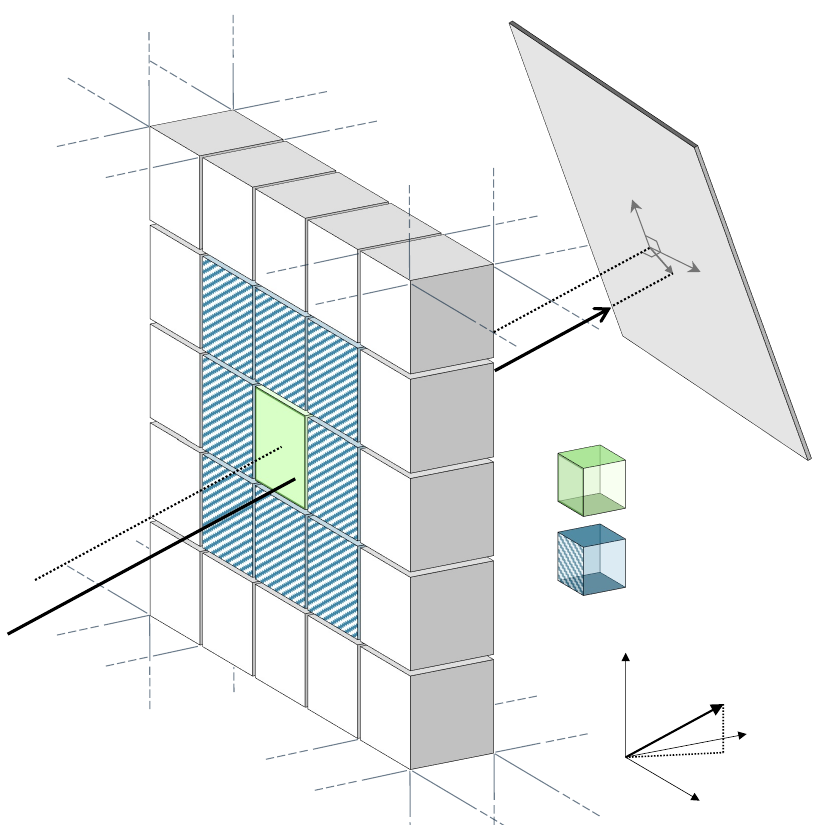}};

  \begin{scope}[x={(img.south east)}, y={(img.north west)}]

    --- Optional helper grid while positioning (turn on as needed) ---
    \node[anchor=west] at (0.75,0.22) {\scriptsize$\be_3$};
    \node[anchor=west] at (0.87,0.098) {\scriptsize\shortstack{$\be_2$\\ \scriptsize(\emph{dominant})}};
    \node[anchor=east] at (0.92,0.025) {\scriptsize$\be_1$};

    \node[anchor=west] at (0.87,0.16) {\scriptsize $\bw$};
    \node[anchor=west] at (0.75,0.425) {\shortstack{\scriptsize Crossed voxel\\ \scriptsize of index $\bk$}};
    \node[anchor=west] at (0.75,0.325) {\shortstack{\scriptsize Neighbor voxels\\ \scriptsize of indices $\bl \in \mathcal{N}_{L}$}};
    \node[anchor=west, rotate=-33] at (0.70,0.955) {\scriptsize \color{gray} Plane $H_{\bw}$};
    \node[anchor=east, rotate=-33] at (0.87,0.675) {\scriptsize\color{gray}$\mathbf{u}$};
    \node[anchor=east, rotate=10] at (0.81,0.76) {\scriptsize\color{gray}$\mathbf{v}$};
    \node[anchor=east] at (0.895,0.65) {\scriptsize\color{gray}$\mathbf{s}_{\bl = \bk}$};



  \end{scope}
\end{tikzpicture}
\caption{Notations for Algorithm~\ref{alg:projection}.}
\label{fig:neighbors}
\end{figure}

\begin{algorithm}[H]
\renewcommand{\baselinestretch}{1.1}\selectfont 
\caption{{SplineSplat} projection algorithm}
\begin{algorithmic}[1]
\Require $(c_\bk)_{\bk \in \Omega}\in\mathbb{R}^{N^3}$, $\bk_0\in\Z^3$, $\bw\in\mathbb{S}^2$, $\bs\in\R^3$, $L\in\R$, trained MLP $f_{\theta^*}$
\State $\bk = \bk_0$ \Comment{first crossed voxel}
\State $P = 0$ \Comment{projection value}
\State $(\neigh^{(-2)}, \neigh^{(-1)}) = (\O, \O)$
\State Compute $(\mathbf{u}, \mathbf{v})$ s.t. $\mathbf{u} \perp \bw$ and $\mathbf{v} = \mathbf{u}\times\bw$
\While {\texttt{DDA} is active}
  \State $\neigh = \texttt{Get\_Neighbors}(\bk, \bw, L, (\neigh^{(-1)}, \neigh^{(-2)}))$
  \State ${\bs}_{\bl} = \text{Proj}_{H_{\bw}}(\bs - \bl), \forall \bl \in \neigh$ 
  \Comment{$H_{\bw} = \mathrm{span}(\mathbf{u}, \mathbf{v})$}
  \State $P \gets P + \sum_{\bl \in \neigh} c_{\bl}~f_{\theta^*}(\bw, {\bs}_{\bl})$
  \State $\bk \gets \texttt{DDA\_Step}(\bk, \bw, \bs)$
  \State $(\neigh^{(-2)}, \neigh^{(-1)}) \gets (\neigh^{(-1)}, \neigh)$
\EndWhile
\State \Return $P$
\end{algorithmic}
\label{alg:projection}
\end{algorithm}
\renewcommand{\baselinestretch}{1.0}\selectfont 
\vspace{-1em}
\begin{algorithm}[H]
\caption{\texttt{Get\_Neighbors}}
\begin{algorithmic}[1]
\Require $\bk\in \Z^3$, $\bw\in\mathbb{S}^2$, $L\in\R$, $(\neigh^{(-1)}, \neigh^{(-2)})$
\State $d^* \in \arg\max_{d=1,2,3} |\omega_d|$ \Comment{dominant axis}
\State $\neigh^{\text{tot}} = \{ \bl\in \mathbb{Z}^3 ~|~ \bl_{d^*} = \bk_{d^*}, \|\bl - \bk\|_\infty \leq \lceil L - \tfrac{1}{2} \rceil \}$
\State \Return  $\neigh = \neigh^{\text{tot}} \setminus (\neigh^{(-1)} \cup \neigh^{(-2)})$
\end{algorithmic}
\label{alg:get_neighbors}
\end{algorithm}

\subsection{Overall Algorithm}

\begin{figure*}[t]
  \centering
  \begin{tikzpicture}[node distance=0.1cm, every node/.style={inner sep=0pt}]
    \node (gt) at (0,0) {\scalebox{1}[-1]{\includegraphics[width=0.28\textwidth]{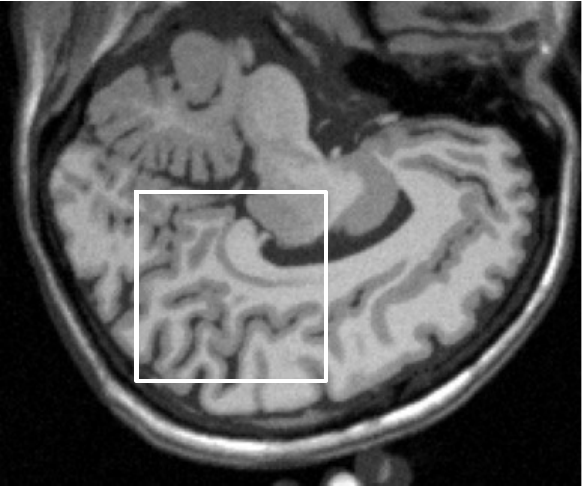}}};
    \node[right=of gt] (spline) {\scalebox{1}[-1]{\includegraphics[width=0.28\textwidth]{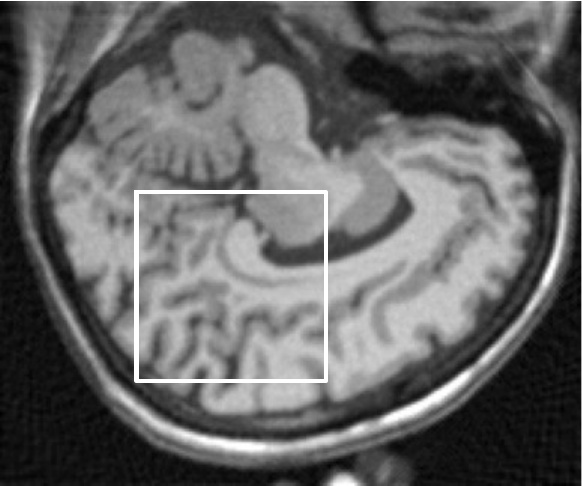}}};
    \node[right=of spline] (pix) {\scalebox{1}[-1]{\includegraphics[width=0.28\textwidth]{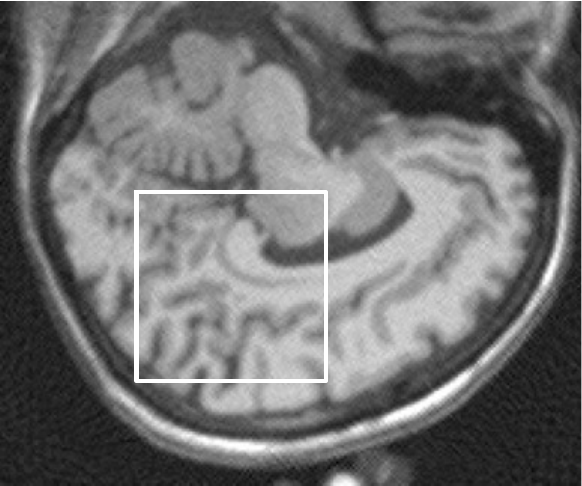}}};
    \node[below=of gt] (gtz2) {\scalebox{1}[-1]{\includegraphics[width=0.28\textwidth]{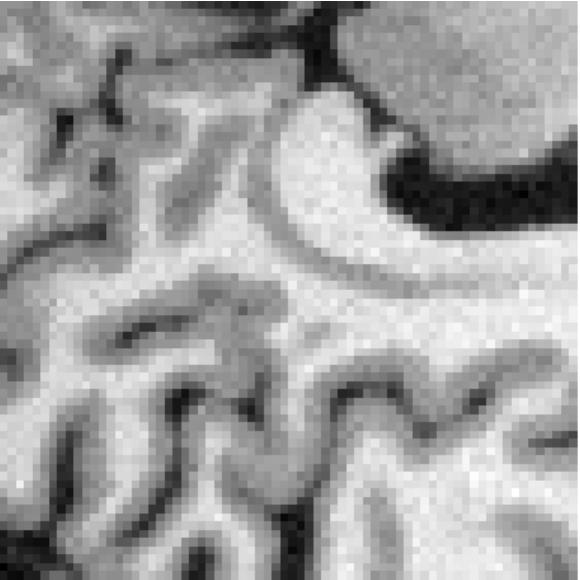}}};
    \node[below=of spline] (splinez2) {\scalebox{1}[-1]{\includegraphics[width=0.28\textwidth]{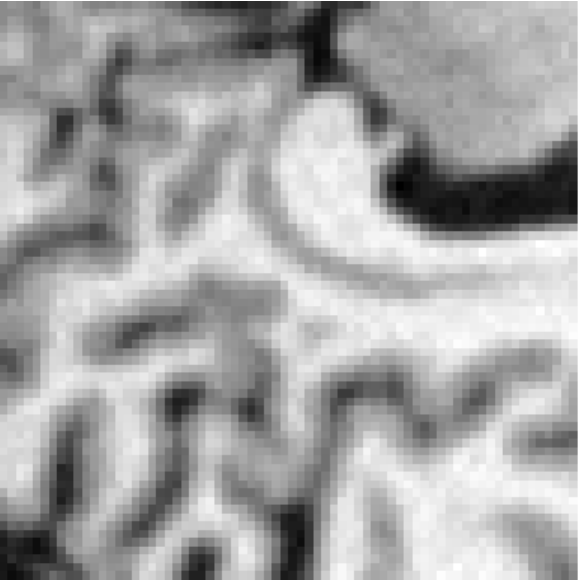}}};
    \node[below=of pix] (pixz2) {\scalebox{1}[-1]{\includegraphics[width=0.28\textwidth]{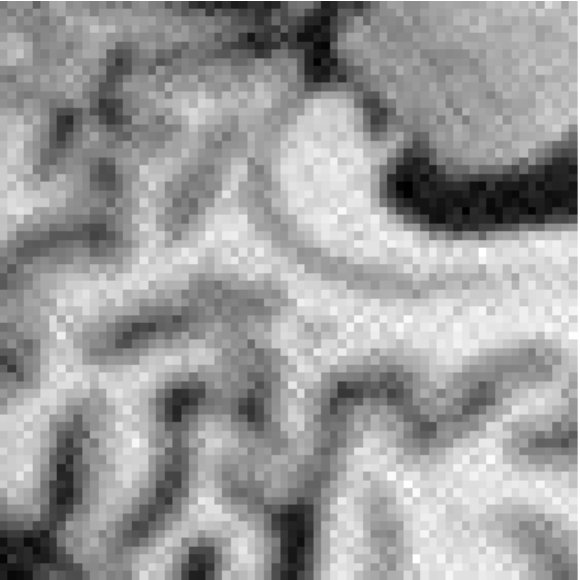}}};
    
    \node[anchor=north east, xshift=-1mm, yshift=-1mm, color=white, font=\small] at (spline.north east) {\shortstack{PSNR\\28.02 dB}};
    \node[anchor=north east, xshift=-1mm, yshift=-1mm, color=white, font=\small] at (pix.north east) {\shortstack{PSNR\\22.29 dB}};
    
    \node[above=0.35cm of gt] {{Reference}};
    \node[above=0.35cm of spline] {{Ours ({SplineSplat})}};
    \node[above=0.35cm of pix] {{Baseline (voxel-based)}};
    \node[left=0.5cm of gt, anchor=center, rotate=90] {Volume slice};
    \node[left=0.5cm of gtz2, anchor=center, rotate=90] {Zoom region};
  \end{tikzpicture}
  \caption{Reconstruction results using our SplineSplat method (middle column) and the baseline method (right column) for a 3D reference of a human brain (left column). Top row: central slice. Bottom row: cutout.}
  \label{fig:reconstructions}
\end{figure*}
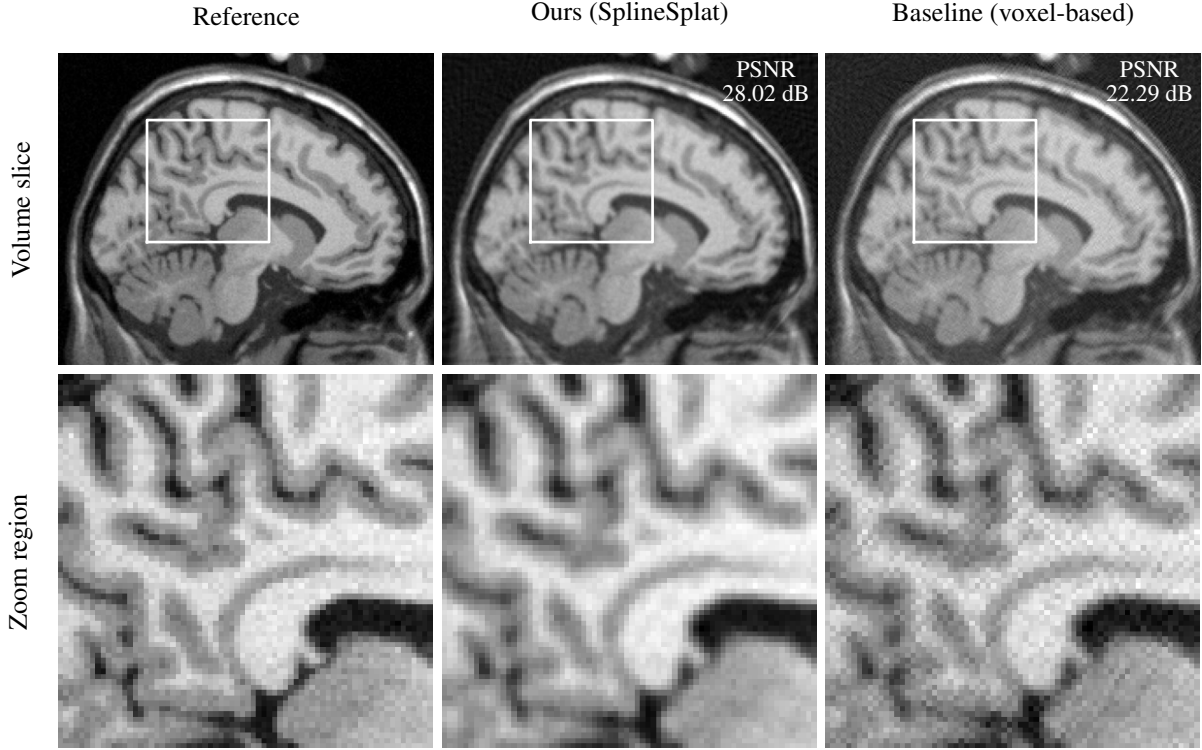
We present our overall projection Algorithm~\ref{alg:projection} and illustrate its main notations in Figure~\ref{fig:neighbors}. It combines the principles described in Section~\ref{sec:principles} with the learned projector of Section~\ref{sec:splinesplat}. At each DDA step, we first let Algorithm~\ref{alg:get_neighbors} find the set of contributing basis functions. We then compute their contributions using the learned projector~$f_{\theta^*}$ and accumulate them to the projection value~$P$. Finally, our modified DDA steps to the next cell while updating the history of the previous neighbors to avoid redundant computations. Algorithm~\ref{alg:projection} with minor changes is applicable to evaluate the adjoint operator $\mathcal{P}^\ast$ as well, which is perfectly matched to the forward projector.
\section{Experiments and results}
We evaluate our method on the 3D Simulated Brain Database\footnote{http://www.bic.mni.mcgill.ca/brainweb/}. We portray in Figure~\ref{fig:reconstructions} a slice of a reference volume along with reconstructions obtained after $50$ iterations of conjugate-gradient-descent that finds the non-regularized least-squares solution
\begin{equation}
  \mathbf{c}^\ast \in \arg\min_{\mathbf{c}\in\R^{N^3}} \|\mathbf{P}_\varphi \mathbf{c} - \mathbf{y}\|_2^2.
\end{equation}
There, $\mathbf{y} \in \R^M$ are noisy projections simulated with the ASTRA~toolbox in a cone-beam geometry corrupted with Gaussian noise of variance $10^{-3}$, $\mathbf{c}\in\R^{N^3}$ are expansion coefficients, and $\mathbf{P}_\varphi \in \R^{M \times N^3}$ is the forward model defined as
\begin{equation}
  [\mathbf{P}_\varphi]_{m,l} = \mathcal{P}\varphi\big(\bw_m, \bs_m - \text{Proj}_{H_{\bw_m}}(\bk_l)\big),
\end{equation}
implemented with Algorithm~\ref{alg:projection}. The data consists of a volume of size $(180 \times 180 \times 50)$ with normalized coefficients. In the considered setup, we acquire $200$ equally spaced views around the first axis with detectors of size $(180\times 50)$.
We compare our method SplineSplat using quadratic B-splines as basis functions against the traditional voxel-based DDA projector. We observe that our method achieves significantly higher reconstruction quality (28.02 dB versus 22.29 dB in PSNR). A similar trend in PSNR gap is observed across the dataset.\footnote{Mean of PSNR gaps: 5.6 dB. Standard deviation: 0.3 dB.}

\section{Conclusion}
We presented a 3D x-ray projection algorithm that extends ray tracing to overlapping, compactly supported basis functions. A shallow neural network efficiently models the basis contributions, which eliminates the need for analytical expressions. Theory and experiments demonstrate that our method with smooth basis functions yields higher reconstruction quality than voxel-based approaches. The proposed method is agnostic to the projection geometry and can be used in various tomographic imaging modalities.

\newpage
\section{Acknowledgements}
The research leading to these results has received funding
from the European Research Council under Grant ERC-
2020-AdG FunLearn-101020573 and by the Swiss National
Science Foundation under Sinergia Grant CRSII5 198569.

\bibliographystyle{IEEEbib}
\bibliography{refs}

@inproceedings{amanitides,
booktitle = {EG 1987-Technical Papers},
editor = {},
title = {{A Fast Voxel Traversal Algorithm for Ray Tracing}},
author = {Amanatides, J. and Woo, A.},
year = {1987},
publisher = {Eurographics Association},
ISSN = {1017-4656},
ISBN = {},
DOI = {10.2312/egtp.19871000}
}

@ARTICLE{astra,
  author    = "van Aarle, W. and Palenstijn, W. J. and Cant, J. and Janssens, E. and Bleichrodt, F. and Dabravolski, A. and De Beenhouwer, J. and Batenburg, K. J. and Sijbers, J.",
  title     = "Fast and flexible X-ray tomography using the {ASTRA} toolbox",
  journal   = "Opt. Express",
  volume    = "24",
  number    = "22",
  pages     = "25129--25147",
  year      = "2016"
}

@ARTICLE{tigre,
  author    = "Biguri, A. and Dosanjh, M. and Hancock, S. and Soleimani, M.",
  title     = "{TIGRE}: a MATLAB-GPU toolbox for {CBCT} image reconstruction",
  journal   = "Biomed. Phys. Eng. Express",
  volume    = "2",
  number    = "5",
  pages     = "055010",
  year      = "2016"
}

@ARTICLE{conrad,
  author    = "Maier, A. and Hofmann, H. G. and Berger, M. and Fischer, P. and Schwemmer, C. and Wu, H. and M{\"u}ller, K. and Hornegger, J. and Choi, J.-H. and Riess, C. and Keil, A. and Fahrig, R.",
  title     = "{CONRAD}: a software framework for cone-beam imaging in radiology",
  journal   = "Med. Phys.",
  volume    = "40",
  number    = "11",
  pages     = "111914",
  year      = "2013"
}

@INPROCEEDINGS{mehrsa_pourya_box_splines,
  author={Pourya, M. and Haouchat, Y. and Unser, M.},
  booktitle={2024 IEEE International Symposium on Biomedical Imaging (ISBI)}, 
  title={A Continuous-Domain Solution for Computed Tomography with Hessian Total-Variation Regularization}, 
  year={2024},
  volume={},
  number={},
  pages={1-5},
  doi={10.1109/ISBI56570.2024.10635730}}

@INPROCEEDINGS{entezari_unser,
  author    = "Entezari, A. and Unser, M.",
  title     = "A Box Spline Calculus for Computed Tomography",
  booktitle = "Proc. {IEEE} Int. Symp. Biomed. Imaging (ISBI)",
  pages     = "600--603",
  year      = "2010"
}

@ARTICLE{haouchat_tci,
  author    = "Haouchat, Y. and Kashani, S. and Th{\'e}venaz, P. and Unser, M.",
  title     = "Generalized Ray Tracing with Basis Functions for Tomographic Projections",
  journal   = "{IEEE} Trans. Comput. Imaging",
  volume    = "11",
  pages     = "1294--1305",
  year      = "2025"
}

@BOOK{natterer_mathstomo,
  author    = "Natterer, F.",
  title     = "The Mathematics of Computerized Tomography",
  publisher = "SIAM",
  address   = "Philadelphia, PA",
  year      = "2001"
}

@ARTICLE{hao_gao,
  author    = "Gao, H.",
  title     = "Fast parallel algorithms for the X-ray transform and its adjoint",
  journal   = "Med. Phys.",
  volume    = "39",
  number    = "11",
  pages     = "7110--7120",
  year      = "2012"
}

@INBOOK{fessler_statistical_reconstruction,
  author    = "Fessler, J. A.",
  title     = "Statistical Image Reconstruction Methods for Transmission Tomography",
  booktitle = "Medical Image Processing and Analysis",
  editor    = "Sonka, M. and Fitzpatric, J. M.",
  pages     = "1--70",
  publisher = "SPIE Press",
  address   = "Bellingham, Washington",
  year      = "2000",
  volume    = "3"
}

@ARTICLE(thevenaz_interpolation,
AUTHOR="Th{\'{e}}venaz, P. and Blu, T. and Unser, M.",
TITLE="Interpolation Revisited",
JOURNAL="{IEEE} Transactions on Medical Imaging",
YEAR="2000",
volume="19",
number="7",
pages="739--758",
month="July",
note="")

@ARTICLE{kingma_adam,
  author    = "Kingma, D. P. and Ba, J.",
  title     = "Adam: A Method for Stochastic Optimization",
  journal   = "CoRR",
  volume    = "abs/1412.6980",
  year      = "2014"
}

@ARTICLE{rattey,
  author={Rattey, P. and Lindgren, A.},
  journal={IEEE Transactions on Acoustics, Speech, and Signal Processing}, 
  title={Sampling the 2-D Radon transform}, 
  year={1981},
  month=oct,
  volume={29},
  number={5},
  pages={994-1002},
  doi={10.1109/TASSP.1981.1163686}}

@article{Brooks1978,
  title = {A new approach to interpolation in computed tomography},
  volume = {2},
  ISSN = {0363-8715},
  url = {http://dx.doi.org/10.1097/00004728-197811000-00010},
  DOI = {10.1097/00004728-197811000-00010},
  number = {5},
  journal = {Journal of Computer Assisted Tomography},
  publisher = {Ovid Technologies (Wolters Kluwer Health)},
  author = {Brooks,  R. A. and Weiss,  G. H. and Talbert,  A. J.},
  year = {1978},
  month = nov,
  pages = {577–585}
}

\end{document}